\documentclass[11pt]{article}

\usepackage[preprint]{acl}
\makeatletter
\@ifpackageloaded{lineno}{%
  \AtBeginDocument{%
    \def\makeLineNumber{%
      \if@firstcolumn\makeLineNumberLeft\else\makeLineNumberRight\fi}%
  }%
}{}
\makeatother

\usepackage{times}
\usepackage{latexsym}

\usepackage[T1]{fontenc}

\usepackage{lmodern}

\pdfmapline{+phvb LMSans10-Bold <lmssbx10.pfb}

\usepackage[utf8]{inputenc}

\usepackage{microtype}

\usepackage{graphicx}
\graphicspath{{figures/}}

\usepackage{booktabs}

\usepackage{amsmath}

\usepackage{enumitem}

\usepackage{multirow}

\usepackage{tikz}
\usetikzlibrary{arrows.meta, positioning, calc}

\title{Can LLMs Judge Better Than They Generate? Evaluating Task Asymmetry, Mechanistic Interpretability and Transferability for In-Context QA}

\author{Sambaran Bandyopadhyay \\
  Adobe Research \\
  \texttt{sambaranb@adobe.com}}

\begin{document}
\maketitle

\begin{abstract}
LLM-as-a-Judge and self-evaluation pipelines implicitly assume that evaluation is easier than generation. We test this in a controlled in-context QA setting where a context passage is the sole information source and each model judges the answer it generated, removing the parametric-knowledge confound of open-domain comparisons. Across four benchmarks (SQuAD 2.0, DROP, HotpotQA, MuSiQue) and two models, evaluation is not uniformly easier: generation accuracy exceeds self-evaluation on three of four, with multi-hop MuSiQue the exception. Attention analysis reveals why: evaluation attends to context 3--5x less than generation does and barely reads the candidate answer. LoRA fine-tuning confirms the asymmetry is not a training artifact: generation fine-tuning induces over-acceptance and evaluation fine-tuning degrades generation. These findings challenge core assumptions in self-evaluation pipelines.
\end{abstract}

\section{Introduction}
\label{sec:intro}

LLMs are now deployed not just as generators of content, but as evaluators of it---a shift with sweeping implications for both research and industry. LLM-as-a-Judge pipelines~\citep{zheng2023judging, liu2023geval} have rapidly become the workhorse of large-scale text evaluation, powering leaderboards, automatic benchmarking, and production quality assessment by substituting for prohibitively costly human annotation~\citep{bavaresco2025llms}. Self-reflection and self-correction frameworks~\citep{asai2024selfrag, bai2022constitutional, huang2024large} use the same model to critique and iteratively revise its own outputs, underpinning the agentic and reasoning systems that have proliferated in recent product releases. Reinforcement learning from human feedback~\citep{ouyang2022training} relies on a reward model---often itself an LLM---to provide the preference signals that align frontier models to human values, foundational to the post-training pipeline of nearly every modern instruction-tuned model. Together, these uses make LLM-based evaluation not a peripheral capability but a load-bearing component of how LLMs are trained, deployed, and trusted. Yet all of them rest on an \emph{implicit foundational assumption}, rarely tested directly: that LLMs can reliably---and often more easily---judge the correctness of an answer than produce one from scratch.

Yet recent empirical work has pushed back. \citet{oh2024generative} find LLMs achieve \emph{lower} accuracy judging generated answers than producing them on TriviaQA; \citet{jiang2025self} show discriminative selection among self-generated candidates is not reliably superior to direct generation; and \citet{lin2025judge} find only weak correlation between generation and judgment ability across 21 tasks. These works share an important confound, however: they operate in the \textbf{open-domain} setting, where both tasks draw on parametric knowledge. A model that has memorized a fact can evaluate answers about it even when free recall fails, attributing any gap to differential memory retrieval rather than to intrinsic task difficulty.

\paragraph{Our contributions.}
We address this gap with a controlled in-context QA framework and three complementary studies. \textbf{First}, we measure the GA--EA gap on four benchmarks spanning extractive, multi-hop, and numerical reasoning, where each model judges the answer it just generated---a direct test of self-evaluation. \textbf{Second}, we probe last-token attention patterns on both tasks to understand \textit{why} the gap exists; to our knowledge, \textit{this is the first application of mechanistic interpretability tools to the generation--evaluation gap.} \textbf{Third}, we fine-tune with LoRA on each task individually and jointly, then evaluate all checkpoints on both tasks, testing whether the parametric structure for generation and evaluation is shared or distinct. 

We scope the study to in-context factual QA with explicit gold answers: correctness is binary and unambiguous, sidestepping the subjectivity of long-form evaluation~\citep{nandy2025language}, and the context passage as the sole information source ensures any GA--EA gap reflects synthesis difficulty versus self-verification rather than differential memory retrieval.

\section{Related Work}
\label{sec:related}

\paragraph{LLMs as evaluators in modern NLP pipelines.}
LLMs are now used as judges, critics, and reward models throughout the model development lifecycle. \textit{LLM-as-a-Judge} frameworks~\citep{zheng2023judging, liu2023geval, bavaresco2025llms} score or rank generated text at scale, replacing costly human annotation. \textit{Self-reflection} and \textit{self-correction} pipelines~\citep{asai2024selfrag, bai2022constitutional, huang2024large} have a model critique and revise its own outputs. LLM-based reward models drive alignment via RLHF~\citep{ouyang2022training}, supplying preference signals for post-training. Known biases such as \textit{self-preference}~\citep{panickssery2024llm} make systematic characterization of LLM evaluation behavior consequential.

\paragraph{Generation--evaluation asymmetry.}
A growing literature compares the same model's generation and evaluation abilities directly. \citet{oh2024generative} find that LLMs perform worse at evaluating their TriviaQA generations than at producing them. \citet{jiang2025self} show that discriminative selection among self-generated candidates is not reliably better than direct generation. \citet{lin2025judge} examine 21 tasks across 11 models and report only weak generation--judgment correlation. All three operate in the open-domain setting where parametric knowledge confounds task difficulty; we complement them by restricting both tasks to a shared in-context source and adding mechanistic and transfer-based analyses.

\paragraph{Mechanistic interpretability of LLM decisions.}
Mechanistic interpretability seeks to explain how transformers route information internally. \citet{elhage2021mathematical} formalize attention heads as read/write circuits; \citet{olsson2022incontext} identify induction heads as a key mechanism for in-context learning. \citet{meng2022rome} and \citet{geva2023ffn} localize factual associations to MLP layers, and \citet{belrose2023logit} show that semantic representations consolidate in later layers of decoder-only transformers. None of these works target the evaluation task or the generation--evaluation gap specifically.

\paragraph{Cross-task transfer with parameter-efficient fine-tuning.}
LoRA~\citep{hu2021lora} adapters isolate the parametric structure each task induces by training on one task and evaluating on another. \citet{dymkiewicz2025donors} document asymmetric donor--recipient transfer patterns across NLP benchmarks. We apply this setup to test whether generation and evaluation share parametric structure within a single QA domain.

\section{Methodology}
\label{sec:method}

\begin{figure}[t]
\centering
\resizebox{\linewidth}{!}{%
\begin{tikzpicture}[
  >=Stealth, thick,
  every node/.style={font=\small},
  inpbox/.style ={draw, fill=blue!8,    rounded corners=4pt,
                  text width=2.6cm, align=center, inner sep=5pt},
  genbox/.style ={draw, fill=orange!18, rounded corners=4pt,
                  text width=3.2cm, align=center, inner sep=5pt},
  orcbox/.style ={draw, fill=teal!13,   rounded corners=4pt,
                  text width=3.2cm, align=center, inner sep=5pt},
  evalbox/.style={draw, fill=orange!18, rounded corners=4pt,
                  text width=3.2cm, align=center, inner sep=5pt},
  mbox/.style   ={draw, dashed, fill=gray!8, rounded corners=3pt,
                  text width=2.9cm, align=center, inner sep=5pt},
  lbl/.style    ={font=\scriptsize\itshape, inner sep=2pt},
]

\node[inpbox] (inp)
  {Input: $(c,\,q,\,a^*)$};

\node[genbox, below=1.0cm of inp] (gen)
  {\textbf{Generator}\;$\mathcal{L}$\\[2pt]
   $\mathcal{T}_{\mathrm{gen}}$:\;$(c,q)\!\to\!a$};

\node[orcbox, below=1.0cm of gen] (orc)
  {\textbf{Oracle}\;$\mathcal{L}^*$\\[2pt]
   $(c,q,a,a^*)\!\to\!y^*$};

\node[evalbox, below=1.0cm of orc] (evl)
  {\textbf{Evaluator}\;$\mathcal{L}$\\[2pt]
   $\mathcal{T}_{\mathrm{eval}}$:\;$(c,q,a)\!\to\!y$};

\node[mbox, right=2.2cm of orc] (ga)
  {\textbf{GA}\\[2pt]$P(y^*\!=\!\textsc{Cor.})$};

\node[mbox, right=2.2cm of evl] (ea)
  {\textbf{EA}\\[2pt]$P(y\!=\!y^*)$};

\node[mbox, below=0.6cm of ea] (delta)
  {$\boldsymbol{\Delta} = \mathrm{EA} - \mathrm{GA}$};


\draw[->]
  (inp) -- node[right, lbl]{$(c,q)$} (gen);

\draw[->]
  (gen) -- node[right, lbl]{$a$} (orc);

\draw[->, dashed]
  (inp.west) -- ++(-0.70,0) |-
  node[left, lbl, pos=0.65]{$a^*$} (orc.west);

\draw[->, dashed]
  (gen.west) -- ++(-1.15,0) |-
  node[left, lbl, pos=0.70]{$a$} (evl.west);

\draw[->]
  (orc.east) -- node[above, lbl]{$y^*$} (ga.west);

\draw[->]
  (evl.east) -- node[above, lbl]{$y$} (ea.west);

\draw[->, dashed]
  (ga.east) -- ++(0.55,0) |-  (delta.east);

\draw[->]
  (ea) -- (delta);

\end{tikzpicture}%
}
\caption{Core task-asymmetry pipeline. Model $\mathcal{L}$ is tested on two tasks per instance: \textbf{generation} ($\mathcal{T}_\text{gen}$) produces answer $a$ from $(c, q)$; \textbf{self-evaluation} ($\mathcal{T}_\text{eval}$) judges whether $a$ is correct given $(c, q, a)$. Oracle $\mathcal{L}^*$ scores $a$ against gold $a^*$, yielding $y^*$ as ground truth for both metrics. Dashed arrows are data-passing operations with no LLM call. $\boldsymbol{\Delta} = \mathrm{EA} - \mathrm{GA}$ is the primary asymmetry measure. The mechanistic and transferability studies extend this pipeline.}
\label{fig:framework}
\end{figure}
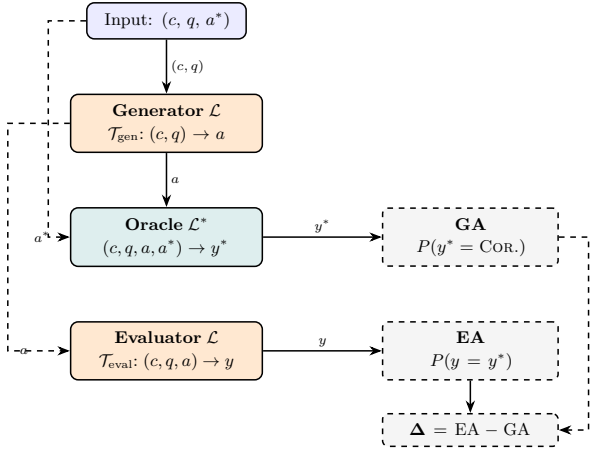

We address the three studies introduced above in Sections~\ref{sec:task_asymmetry}--\ref{sec:transferability}. Figure~\ref{fig:framework} sketches the shared pipeline they build on.

\subsection{Task Asymmetry Analysis}
\label{sec:task_asymmetry}

Each benchmark instance is a triple $(c, q, a^*)$, where $c$ is a context passage, $q$ is a question, and $a^*$ is the gold answer synthesized directly from $c$. We measure a language model $\mathcal{L}$ on two sequential tasks over the same instances.

\paragraph{Generation task ($\mathcal{T}_\text{gen}$).}
Given $(c, q)$, $\mathcal{L}$ produces a free-text answer:
\begin{equation}
    a = \mathcal{L}(c,\, q).
\end{equation}

\paragraph{Evaluation task ($\mathcal{T}_\text{eval}$).}
The generated answer $a$ from $\mathcal{T}_\text{gen}$ is passed directly to $\mathcal{L}$ as a candidate to be judged. Given $(c, q, a)$, $\mathcal{L}$ produces a binary judgment:
\begin{equation}
    y = \mathcal{L}(c,\, q,\, a) \;\in\; \{\textsc{Correct},\, \textsc{Incorrect}\}.
\end{equation}

This sequential design makes $\mathcal{T}_\text{eval}$ a \textit{self-evaluation} task: $\mathcal{L}$ judges the quality of an answer it has just produced. The same sampled instances are used for both tasks, ensuring direct comparability. We summarize the resulting asymmetry by $\Delta = \mathrm{EA} - \mathrm{GA}$, with EA and GA formally defined in Section~\ref{sec:metrics}.

\paragraph{Oracle scoring.}
We use an oracle LLM $\mathcal{L}^*$ to score the generated answer $a$ against the gold answer $a^*$:
\begin{equation}
\begin{split}
    y^* &= \mathcal{L}^*(c, q, a, a^*) \\
        &\in \{\textsc{Correct}, \textsc{Incorrect}\}.
\end{split}
\end{equation}
$y^*$ serves a dual purpose: it defines \textit{generation accuracy} (whether $a$ is correct) and provides the \textit{ground-truth evaluation label} against which $\mathcal{L}$'s judgment $y$ is compared. Using a single oracle signal for both tasks ensures GA and EA are grounded in a consistent external reference.

\paragraph{Prompting.}
Both tasks use zero-shot prompting with fixed templates. The generation prompt instructs $\mathcal{L}$ to answer using only the provided context, responding concisely. The evaluation prompt presents $(c, q, a)$ and instructs $\mathcal{L}$ to judge whether $a$ is correct given the context, responding with exactly \textit{``Correct''} or \textit{``Incorrect.''} Identical context formatting is used across both tasks.

\subsection{Mechanistic Interpretation}
\label{sec:mech_interp}

To understand the mechanism behind any observed gap, we examine how $\mathcal{L}$ internally allocates attention on the two tasks. For each prompt we perform a single forward pass with attention outputs enabled and extract the \textit{last-token attention row}---the distribution over all input positions the model uses when predicting the first output token. This is the natural decision point for both tasks, since $\mathcal{T}_\text{gen}$ and $\mathcal{T}_\text{eval}$ both produce short outputs (typically 1--3 tokens for $\mathcal{T}_\text{gen}$ gold answers and a single token for $\mathcal{T}_\text{eval}$), making the first-token decision the most consequential one.

We measure the fraction of attention mass falling on three labeled prompt spans---context $c$, question $q$, and candidate answer $a$ (the latter only in $\mathcal{T}_\text{eval}$)---yielding a span-level fingerprint of each task's allocation strategy. Attention is averaged over all heads and over the final eight transformer layers (layers~24--31 of the 32-layer Llama-3.1-8B-Instruct). Late layers in decoder-only transformers consolidate high-level semantic and task-relevant representations, while earlier layers focus on lexical and positional features~\citep{belrose2023logit, geva2023ffn}; we therefore restrict the ratio computation to this band to isolate the semantic, decision-relevant attention pattern from low-level token-matching behavior.

\subsection{Transferability Analysis}
\label{sec:transferability}

To test whether the parametric structure used for generation and evaluation is shared, we fine-tune $\mathcal{L}$ with LoRA adapters in three configurations:
\begin{itemize}[noitemsep, topsep=2pt]
    \item \textbf{LoRA-Gen}: trained only on $(c, q) \to a^*$ examples.
    \item \textbf{LoRA-Eval}: trained only on $(c, q, a) \to y^*$ examples, with $a$ drawn either from a hard-negative pool (described below) or, when unavailable, from answer rotation.
    \item \textbf{LoRA-Both}: trained on a balanced 1:1 union of the two task formats.
\end{itemize}
Each adapter is then evaluated on \emph{both} $\mathcal{T}_\text{gen}$ and $\mathcal{T}_\text{eval}$ using the same prompting templates and oracle scoring described in Section~\ref{sec:task_asymmetry}. Asymmetric cross-task gains would indicate that the two abilities draw on partially distinct parametric resources; symmetric gains would indicate a shared underlying capability.

\paragraph{Hard-negative generation for evaluator training.}
A naive negative for $\mathcal{T}_\text{eval}$ training---an arbitrarily picked wrong answer---is trivially distinguishable from a correct one and produces an evaluator that does not generalize to plausible mistakes. We instead generate \textit{plausible but incorrect} candidates by querying $\mathcal{L}$ \emph{without} the context passage, forcing it to rely on parametric memory and produce hallucinated answers. Up to five hallucinations are attempted per record; the first whose answer does \emph{not} match the gold $a^*$ (judged by the same oracle $\mathcal{L}^*$) is kept as the hard negative for that record. Refusal-style outputs (e.g.\ ``I need more information'') are filtered, since they teach the evaluator to detect non-answers rather than to verify factual content; records that yield no usable hard negative fall back to answer rotation, where $a$ is sampled from another instance's gold answer. The resulting training set thus mirrors the distribution of mistakes the evaluator will face at test time, where it judges actual model-generated answers.

\section{Experimental Setup}
\label{sec:experiments}

\subsection{Datasets}

We use the validation split of each dataset exclusively for the task-asymmetry and attention analyses, and the training split only for LoRA fine-tuning. A detailed summary of dataset usage---including per-experiment sample counts and the rationale for downsampling---is given in Appendix~\ref{sec:dataset_usage} (Table~\ref{tab:datasets_extended}).

\textbf{SQuAD~2.0}~\citep{rajpurkar2018know} contains extractive reading comprehension over Wikipedia paragraphs; we retain only answerable questions, yielding 5,928 instances. Both tasks make minimal synthesis demands, providing our most extractive benchmark.

\textbf{DROP}~\citep{dua2019drop} tests discrete numerical reasoning over passage-length texts; we retain number-type answers, yielding 5,889 instances. Generating the correct number requires multi-step arithmetic; evaluating it reduces to numeric comparison.

\textbf{HotpotQA}~\citep{yang2018hotpotqa} requires 2-hop reasoning across two supporting Wikipedia passages plus eight distractors (all provided); we retain non-yes/no questions, yielding 6,947 instances.

\textbf{MuSiQue}~\citep{trivedi2022musique} is a multi-hop benchmark with annotated hop counts (1,252 $\times$ 2-hop; 760 $\times$ 3-hop; 405 $\times$ 4-hop), allowing us to stratify the gap by reasoning depth. We use the full answerable validation split of 2,417 instances, with 20 paragraphs provided per question.

\subsection{Models}

We evaluate two models as $\mathcal{L}$: \texttt{Llama-3.1-8B-Instruct}~\citep{dubey2024llama}, a capable open-source model at modest scale, and \texttt{GPT-4o-mini}, a stronger proprietary model. Comparing these two lets us examine whether the generation--evaluation gap varies with capability. We use \texttt{GPT-4o} as the oracle $\mathcal{L}^*$ throughout~\citep{zheng2023judging}. For the attention analysis, we use only the open-source Llama-3.1-8B-Instruct, as it provides the internal attention weights required by the analysis.

\subsection{Implementation Details}

All models are queried at temperature $\tau = 0$. We sample 500 instances uniformly from each validation split. Each instance is processed sequentially: $\mathcal{L}$ generates answer $a$, the oracle assigns $y^*$, and $\mathcal{L}$ evaluates the same $a$. Responses that do not begin with \textsc{Correct} or \textsc{Incorrect} are treated as abstentions and excluded from metric computation.

For the attention analysis, we expose raw attention weights via a single forward pass per prompt and extract the last-token attention row. Span boundaries are identified via character-to-token offset mapping. Analysis uses the 184 jointly verified samples (Section~\ref{sec:oracle_reliability}).

For the transferability study, we fine-tune Llama-3.1-8B-Instruct with LoRA (rank 16, $\alpha = 32$) on the query, key, value, and output projections, training on 5,000 instances per dataset. Full hyperparameters and hardware details are in Appendix~\ref{sec:reproducibility}.

\subsection{Metrics}
\label{sec:metrics}

For each instance, we have the oracle label $y^*$ and $\mathcal{L}$'s judgment $y$, from which we compute:

\textbf{Generation Accuracy (GA)}: $\mathrm{GA} = \frac{1}{n}\sum_i \mathbf{1}[y^*_i = \textsc{Correct}]$, the fraction of generated answers judged correct by the oracle.

\textbf{Evaluation Accuracy (EA)}: $\mathrm{EA} = \frac{1}{n}\sum_i \mathbf{1}[y_i = y^*_i]$, the fraction where $\mathcal{L}$'s self-judgment matches the oracle. \textbf{Gap} $\boldsymbol{\Delta} = \mathrm{EA} - \mathrm{GA}$ is our primary asymmetry measure (positive $\Rightarrow$ self-evaluation exceeds generation).

\textbf{Evaluation Precision, Recall, and F1 (EP / ER / EF1)} characterize $\mathcal{L}$ as a binary classifier with \textsc{Correct} as the positive class, and are used to diagnose over-acceptance bias~\citep{liu2023geval}.

\section{Results and Discussion}
\label{sec:discussion}

We organize our findings around five research questions, beginning with the reliability of the oracle that grounds GA and EA.

\subsection{R1: Is the Oracle LLM Reliable?}
\label{sec:oracle_reliability}

Before turning to the substantive questions, we validate the oracle $\mathcal{L}^*$ that grounds both GA and EA. A concern with any LLM-as-oracle framework is that the oracle's judgments may themselves be unreliable. In our setting this concern is mitigated by the nature of the task: answers on all four benchmarks are short factual responses---typically a single word, number, or key phrase---paired with an explicit gold answer $a^*$, making the oracle's task closer to fuzzy string matching than open-ended evaluation.

To quantify this empirically, we randomly sampled 50 instances per dataset (200 total) from the Llama-3.1-8B-Instruct results and presented each $(c, q, a, a^*)$ tuple to two judges independently: GPT-4o (our main oracle) and GPT-5.4 (a stronger, more recent model, used as a super-oracle reference). Table~\ref{tab:oracle_reliability} reports inter-model agreement.

\begin{table}[t]
\centering
\small
\begin{tabular}{lrrrr}
\toprule
\textbf{Dataset} & \textbf{N} & \textbf{Agree\%} & \textbf{FP} & \textbf{FN} \\
\midrule
SQuAD~2.0 & 50  & 98.0 & 1  & 0 \\
DROP       & 50  & 92.0 & 2  & 2 \\
HotpotQA  & 50  & 88.0 & 3  & 3 \\
MuSiQue   & 50  & 90.0 & 5  & 0 \\
\midrule
Overall   & 200 & 92.0 & 11 & 5 \\
\bottomrule
\end{tabular}
\caption{GPT-4o vs.\ GPT-5.4 oracle agreement on 50 sampled Llama-3.1-8B-Instruct outputs per dataset. FP = GPT-4o \textsc{Cor.}, GPT-5.4 \textsc{Inc.}; FN = opposite. GPT-4o intra-model consistency (fresh vs.\ stored call) is 99.5\% across all 200 samples.}
\label{tab:oracle_reliability}
\end{table}

GPT-4o and GPT-5.4 agree on 92\% of judgments overall, with near-perfect agreement on extractive SQuAD~2.0 (98\%) and no systematic directional bias across the 16 disagreements. GPT-4o's re-call consistency is 99.5\%. We therefore treat GPT-4o as a reliable oracle and use the 184 jointly verified samples (where both judges agree) as the basis for the attention analysis.

\subsection{R2: Is Self-Evaluation Easier Than Generation?}
\label{sec:r1}

\begin{table*}[t]
\centering
\small
\begin{tabular}{lrrrrrrrrrrrrr}
\toprule
& \multicolumn{3}{c}{\textbf{SQuAD~2.0}} & \multicolumn{3}{c}{\textbf{DROP}} & \multicolumn{3}{c}{\textbf{HotpotQA}} & \multicolumn{3}{c}{\textbf{MuSiQue}} \\
\cmidrule(lr){2-4}\cmidrule(lr){5-7}\cmidrule(lr){8-10}\cmidrule(lr){11-13}
\textbf{Model} & GA & EA & $\Delta$ & GA & EA & $\Delta$ & GA & EA & $\Delta$ & GA & EA & $\Delta$ \\
\midrule
Llama-3.1-8B-Instruct & 95.6 & 92.2 & $-3.4$ & 63.4 & 62.4 & $-1.0$  & 83.2 & 69.0 & $-14.2$ & 55.6 & 59.2 & $+3.6$ \\
GPT-4o-mini           & 97.8 & 96.6 & $-1.2$ & 79.6 & 73.8 & $-5.8$  & 88.0 & 86.0 & $-2.0$  & 64.8 & 68.8 & $+4.0$ \\
\bottomrule
\end{tabular}
\caption{Generation Accuracy (GA) and Evaluation Accuracy (EA) (\%) per model--dataset pair. $\Delta = \mathrm{EA} - \mathrm{GA}$; positive indicates self-evaluation exceeds generation.}
\label{tab:main_results}
\end{table*}

Table~\ref{tab:main_results} reports GA, EA and $\Delta$ across all model--dataset pairs. \textbf{The answer to our titular question is: not in general.} $\Delta < 0$ on three of four benchmarks for both models, indicating that generation accuracy \emph{exceeds} self-evaluation accuracy in the majority of settings. The lone exception is MuSiQue, where $\Delta = +3.6$ (Llama) and $+4.0$ (GPT-4o-mini).

\paragraph{Task-type dependence.}
The pre-experiment intuition---that verifying a numeric answer reduces to exact-match checking---is sharply overturned on DROP: $\Delta = -1.0$ (Llama) and $-5.8$ (GPT-4o-mini), indicating that numeric self-evaluation is \emph{harder}, not easier, than generation. A plausible explanation is that a model relying on a faulty arithmetic procedure to produce an answer cannot detect the resulting error during evaluation, because the same flawed procedure underlies its verification attempt. The largest negative gap overall belongs to HotpotQA for Llama ($-14.2$); the presence of the candidate answer $a$ in the evaluation prompt appears to disrupt the model's ability to independently trace the 2-hop reasoning chain needed to assess it. SQuAD~2.0 shows the smallest absolute gaps for both models, consistent with minimal synthesis demands in the extractive setting.

\begin{table*}[t]
\centering
\resizebox{\textwidth}{!}{%
\begin{tabular}{lrrrrrrrrrrrr}
\toprule
& \multicolumn{3}{c}{\textbf{SQuAD~2.0}} & \multicolumn{3}{c}{\textbf{DROP}} & \multicolumn{3}{c}{\textbf{HotpotQA}} & \multicolumn{3}{c}{\textbf{MuSiQue}} \\
\cmidrule(lr){2-4}\cmidrule(lr){5-7}\cmidrule(lr){8-10}\cmidrule(lr){11-13}
\textbf{Model} & EP & ER & EF1 & EP & ER & EF1 & EP & ER & EF1 & EP & ER & EF1 \\
\midrule
Llama-3.1-8B-Instruct & 97.0 & 94.8 & 95.9 & 74.5 & 61.8 & 67.6 & 87.4 & 73.3 & 79.7 & 67.5 & 51.4 & 58.4 \\
GPT-4o-mini           & 98.0 & 98.6 & 98.3 & 81.9 & 86.2 & 84.0 & 90.9 & 93.4 & 92.2 & 70.8 & 88.3 & 78.6 \\
\bottomrule
\end{tabular}}
\caption{Full evaluation breakdown: Precision (EP), Recall (ER), and F1 (EF1, \%) per model--dataset pair.}
\label{tab:ep_er}
\end{table*}

\paragraph{Acquiescence bias.}
The EP/ER decomposition (Table~\ref{tab:ep_er}) reveals opposite biases: Llama exhibits EP $>$ ER on all four datasets---most pronounced on DROP ($74.5 / 61.8$) and MuSiQue ($67.5 / 51.4$)---indicating a \emph{conservative bias} that over-predicts \textsc{Incorrect} and directly explains its severe negative $\Delta$ on HotpotQA. GPT-4o-mini shows the inverse on three of four datasets, most pronounced on MuSiQue (ER $= 88.3$\%, EP $= 70.8$\%)---a mild over-acceptance bias that inflates EA there and partially drives $\Delta > 0$. Counter-intuitively, the stronger model is the more over-accepting evaluator.

\subsection{R3: How Does Reasoning Complexity Modulate the Gap?}
\label{sec:r2}

MuSiQue's per-question hop annotations let us stratify the gap by reasoning depth within a single dataset---an analysis impossible with HotpotQA, which is uniformly 2-hop.

\begin{table}[t]
\centering
\resizebox{\linewidth}{!}{%
\begin{tabular}{llrrrr}
\toprule
\textbf{Model} & \textbf{Hops} & \textbf{\#Samples} & GA & EA & $\Delta$ \\
\midrule
\multirow{3}{*}{Llama-3.1-8B}
  & 2-hop & 282 & 57.8 & 56.7 & $-1.1$ \\
  & 3-hop & 140 & 55.0 & 65.7 & $+10.7$ \\
  & 4-hop &  78 & 48.7 & 56.4 & $+7.7$ \\
\midrule
\multirow{3}{*}{GPT-4o-mini}
  & 2-hop & 282 & 67.4 & 72.0 & $+4.6$ \\
  & 3-hop & 140 & 66.4 & 68.6 & $+2.1$ \\
  & 4-hop &  78 & 52.6 & 57.7 & $+5.1$ \\
\bottomrule
\end{tabular}}
\caption{Generation--evaluation gap on MuSiQue stratified by hop count. The sharp $\Delta$ increase from 2 to 3 hops for Llama reflects generation quality degrading faster than evaluation accuracy as depth grows.}
\label{tab:hop_analysis}
\end{table}

Table~\ref{tab:hop_analysis} shows the evolution of $\Delta$ across hop counts. For Llama, $\Delta$ transitions sharply from $-1.1$ at 2 hops to $+10.7$ at 3 hops and $+7.7$ at 4 hops: generation accuracy falls to 55.0\% and 48.7\% at 3 and 4 hops respectively, while evaluation accuracy remains comparatively stable (65.7\%, 56.4\%). This asymmetry is consistent with a \emph{candidate-answer-insulation} hypothesis: when $a$ is already present in the evaluation prompt, the model need not reconstruct the full reasoning chain, gaining a partial advantage over generation at high depth. GPT-4o-mini exhibits a consistently positive $\Delta$ at all hop counts ($+4.6$, $+2.1$, $+5.1$), with less variation, reflecting its greater robustness on both tasks. The modest compression at 3 hops for both models suggests a qualitative threshold between 2-hop and higher-order reasoning rather than smooth linear degradation.

\subsection{R4: What Internal Mechanisms Underlie the Gap?}
\label{sec:r3}

To probe the mechanism behind the observed gap, we analyze last-token attention patterns of Llama-3.1-8B-Instruct on $\mathcal{T}_\text{gen}$ and $\mathcal{T}_\text{eval}$ across the 184 jointly verified samples (Section~\ref{sec:oracle_reliability}). The single forward pass per prompt extracts the attention distribution that the model assigns over input positions when predicting the first output token. Attention is averaged over all heads and over layers 24--31. We measure the fraction of attention mass directed at three labeled spans: context $c$, question $q$, and candidate answer $a$ ($\mathcal{T}_\text{eval}$ only). The remaining mass---typically 80--98\%---is absorbed by tokens outside these spans (system prompt, chat template markers, instruction text).

\begin{figure}[t]
\centering
\includegraphics[width=\linewidth]{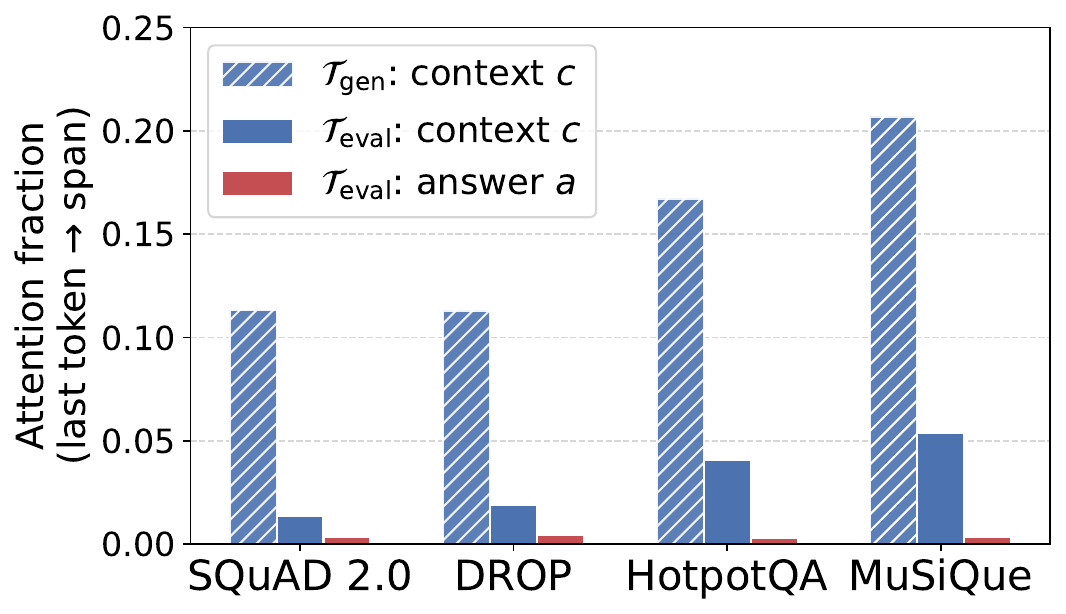}
\caption{Mean last-token attention fraction directed to context ($c$) and candidate answer ($a$) for $\mathcal{T}_\text{gen}$ and $\mathcal{T}_\text{eval}$, averaged over layers 24--31 and 184 jointly verified Llama-3.1-8B-Instruct samples. $\mathcal{T}_\text{eval}$ consistently de-attends to context by 3--5$\times$ relative to $\mathcal{T}_\text{gen}$ and allocates negligible attention (0.3--0.5\%) to the candidate answer it is judging.}
\label{fig:attention_ratio}
\end{figure}

Figure~\ref{fig:attention_ratio} reports the mean attention fractions. $\mathcal{T}_\text{gen}$ directs 11.3--20.7\% of last-token attention to the context passage, with the fraction growing with task complexity (SQuAD~2.0/DROP: 11.3\% $<$ HotpotQA: 16.7\% $<$ MuSiQue: 20.7\%), consistent with multi-hop generation requiring deeper context engagement. $\mathcal{T}_\text{eval}$, by contrast, allocates only 1.4--5.4\% of attention to context---a 3--5$\times$ reduction across all four datasets. More strikingly, $\mathcal{T}_\text{eval}$ devotes only 0.3--0.5\% of attention to the candidate answer $a$ it is tasked with judging.

\paragraph{Interpretation.}
The consistent context de-attention in $\mathcal{T}_\text{eval}$ is a \emph{structural signature} of the evaluation task: rather than re-reading the passage to verify the answer, the model's last-token decision relies primarily on structural and instructional tokens. This pattern directly explains why $\Delta$ is negative on three of the four benchmarks. On SQuAD~2.0, DROP, and HotpotQA, $\mathcal{T}_\text{gen}$ allocates 11.3--16.7\% of last-token attention to the context while $\mathcal{T}_\text{eval}$ allocates only 1.4--4.1\%---insufficient to re-trace the lookup $\mathcal{T}_\text{gen}$ performs at synthesis time. The gap widens precisely where context engagement matters most: HotpotQA's largest negative $\Delta = -14.2$ (Llama) coincides with the largest gen-to-eval reduction in context attention (16.7\%~$\to$~4.1\%), and SQuAD~2.0's smaller $\Delta = -3.4$ matches the smallest such reduction (11.3\%~$\to$~1.4\%) on a task whose synthesis demands are minimal. The lone exception is MuSiQue, where $\Delta > 0$ despite an even sharper context-attention drop ($\Delta_\text{ctx} = -0.153$); the attention pattern alone is therefore not sufficient to explain this inversion, and we defer a fuller discussion of it to the Limitations section.

\paragraph{Extended analyses.}
Appendix~\ref{sec:extended_attention} extends this analysis along three orthogonal axes---paragraph-level selectivity (supporting vs.\ distractor), answer-mention concentration within the context, and causal ablations of the candidate-answer slot---which together sharpen the mechanistic picture sketched above.

\begin{table*}[!t]
  \centering
  \resizebox{\textwidth}{!}{%
  \begin{tabular}{lrrrrrrrrrrrrr}
  \toprule
  & \multicolumn{3}{c}{\textbf{SQuAD~2.0}} & \multicolumn{3}{c}{\textbf{DROP}} & \multicolumn{3}{c}{\textbf{HotpotQA}} & \multicolumn{3}{c}{\textbf{MuSiQue}} \\
  \cmidrule(lr){2-4}\cmidrule(lr){5-7}\cmidrule(lr){8-10}\cmidrule(lr){11-13}
  \textbf{Checkpoint} & GA & EA & $\Delta$ & GA & EA & $\Delta$ & GA & EA & $\Delta$ & GA & EA & $\Delta$ \\
  \midrule
  Base (Llama-3.1-8B)          & 95.6 & 92.2 & $-3.4$  & 63.4 & 62.4 & $-1.0$  & 83.2 & 69.0 & $-14.2$ & 55.6 & 59.2 & $+3.6$  \\
  \;\;+ LoRA-Gen (joint)       & 94.0 & 94.6 & $+0.6$  & 61.2 & 62.0 & $+0.8$  & 84.0 & 84.4 & $+0.4$  & 49.0 & 56.8 & $+7.8$  \\
  \;\;\;\;+ LoRA-Gen (per-ds)  & 92.4 & 93.2 & $+0.8$  & 64.0 & 65.2 & $+1.2$  & 83.4 & 84.2 & $+0.8$  & 48.4 & 53.1 & $+4.7$  \\
  \;\;+ LoRA-Eval (joint)      & 94.4 & 78.4 & $-16.0$ & 46.6 & 57.8 & $+11.2$ & 76.6 & 63.6 & $-13.0$ & 38.0 & 60.8 & $+22.8$ \\
  \;\;\;\;+ LoRA-Eval (per-ds) & 90.6 & 88.4 & $-2.2$  & 41.4 & 61.2 & $+19.8$ & 77.2 & 47.5 & $-29.7$ & 36.8 & 59.6 & $+22.8$ \\
  \;\;+ LoRA-Both (joint)      & 92.8 & 93.2 & $+0.4$  & 58.8 & 61.4 & $+2.6$  & 82.4 & 83.0 & $+0.6$  & 49.4 & 58.4 & $+9.0$  \\
  \;\;\;\;+ LoRA-Both (per-ds) & 92.2 & 92.0 & $-0.2$  & 62.5 & 61.7 & $-0.8$  & 83.2 & 83.2 & $\pm0.0$  & 44.3 & 53.5 & $+9.2$  \\
  \bottomrule
  \end{tabular}
  }
  \caption{GA, EA, and $\Delta$ (\%) under LoRA fine-tuning. Joint trains on all datasets; per-ds on one at a time. EA gains under LoRA-Gen/Both are acquiescence-driven (Table~\ref{tab:transfer_eper}), not genuine transfer.}
  \label{tab:transfer_main}
  \end{table*}
  
  \begin{table*}[!t]
  \centering
  \resizebox{0.9\textwidth}{!}{%
  \begin{tabular}{lrrrrrrrrrrrr}
  \toprule
  & \multicolumn{3}{c}{\textbf{SQuAD~2.0}} & \multicolumn{3}{c}{\textbf{DROP}} & \multicolumn{3}{c}{\textbf{HotpotQA}} & \multicolumn{3}{c}{\textbf{MuSiQue}} \\
  \cmidrule(lr){2-4}\cmidrule(lr){5-7}\cmidrule(lr){8-10}\cmidrule(lr){11-13}
  \textbf{Checkpoint} & EP & ER & EF1 & EP & ER & EF1 & EP & ER & EF1 & EP & ER & EF1 \\
  \midrule
  Base (Llama-3.1-8B)       & 97.0 & 94.8 & 95.9 & 74.5 & 61.8 & 67.6 & 87.4 & 73.3 & 79.7 & 67.5 & 51.4 & 58.4 \\
  \;\;+ LoRA-Gen (joint)    & 94.9 & 99.6 & 97.2 & 62.5 & 94.8 & 75.3 & 84.3 & 100.0 & 91.5 & 56.3 & 100.0 & 72.0 \\
  \;\;+ LoRA-Eval (joint)   & 95.3 & 81.1 & 87.6 & 57.9 & 34.8 & 43.4 & 81.5 & 67.9 & 74.1 & 54.0 & 58.8 & 56.3 \\
  \;\;+ LoRA-Both (joint)   & 93.4 & 99.8 & 96.5 & 60.7 & 97.3 & 74.8 & 83.3 & 99.3 & 90.6 & 57.5 & 99.2 & 72.8 \\
  \bottomrule
  \end{tabular}}
  \caption{EP, ER, and EF1 (\%) per checkpoint, \textsc{Correct} as positive. LoRA-Gen/Both drive ER\,${\to}\,100\%$ (acquiescence); LoRA-Eval drives it too low. MuSiQue abstention: 13.0\% (Gen/Both), 15.4\% (Eval).}
  \label{tab:transfer_eper}
  \end{table*}
  
  \begin{table}[!ht]
  \centering
  \small
  \begin{tabular}{lrr}
  \toprule
  \textbf{Checkpoint} & $\Delta$GA (avg) & $\Delta$EA (avg) \\
  \midrule
  LoRA-Gen (joint)   & $-2.4$  & $+3.7$ \\
  LoRA-Eval (joint)  & $-10.6$ & $-5.6$ \\
  LoRA-Both (joint)  & $-3.6$  & $+3.3$ \\
  \bottomrule
  \end{tabular}
  \caption{Average gain over base across four datasets (pp), joint adapters only.}
  \label{tab:transfer_gain}
  \end{table}

\subsection{R5: Does Cross-Task Transfer Occur Under LoRA Fine-Tuning?}
\label{sec:r4}

To test whether the parametric structure underlying generation and evaluation is shared, we fine-tune Llama-3.1-8B-Instruct with LoRA in three configurations (LoRA-Gen, LoRA-Eval, LoRA-Both; see Section~\ref{sec:transferability}) and evaluate each on both tasks across all four datasets.

\paragraph{Hypotheses.}
We pre-register three expectations: (i) LoRA-Gen should raise GA; whether EA follows tests generation-to-evaluation transfer. (ii) LoRA-Eval should raise EA; whether GA improves tests the reverse. (iii) LoRA-Both tests whether joint training delivers both gains or one ability dominates the parameter budget. Asymmetric outcomes would indicate the two abilities draw on overlapping but not identical parametric structure~\citep{dymkiewicz2025donors}.

\paragraph{LoRA-Gen flips $\Delta$ on every dataset---but via acquiescence.}
After LoRA-Gen, $\Delta \geq 0$ on all four datasets (Table~\ref{tab:transfer_main}), with the largest swing on HotpotQA ($-14.2 \to +0.4$, driven by EA rising from $69.0$ to $84.4\%$). However, the EP/ER decomposition (Table~\ref{tab:transfer_eper}) reveals this is entirely a recall-side bias shift: evaluator recall reaches $100\%$ on HotpotQA and MuSiQue, $99.6\%$ on SQuAD~2.0, and $94.8\%$ on DROP---the adapters predict \textsc{Correct} for nearly every input. LoRA-Both shows the same acquiescence pattern ($\mathrm{ER} \geq 97\%$ on every dataset). EA rises only because the underlying GA is high enough that near-uniform \textsc{Correct} responses happen to match the oracle most of the time. This is the acquiescence failure mode flagged in Section~\ref{sec:r1}, driven to an extreme by training.

\paragraph{LoRA-Eval degrades both tasks.}
Training only on the evaluation task is the most striking failure mode. GA falls by $16.8$\,pp on DROP, $17.6$\,pp on MuSiQue, and $6.6$\,pp on HotpotQA, even though the adapter never saw a generation objective. EA also drops on SQuAD~2.0 and HotpotQA. The EP/ER pattern is the inverse of LoRA-Gen: ER drops to $34.8\%$ on DROP and $58.8\%$ on MuSiQue, making the model overly conservative. LoRA-Eval as configured is not a usable adapter.

\paragraph{The MuSiQue $\Delta > 0$ inversion is structural.}
MuSiQue is the only dataset where the base model had $\Delta > 0$. The inversion persists across all three adapters at progressively larger magnitudes ($+7.8$, $+22.8$, $+9.0$). The invariance across very different training regimes directly evidences that this reflects the underlying task pair---multi-hop generation is intrinsically harder than verifying a presented answer---rather than a base-model artifact, supporting the candidate-answer-insulation hypothesis of Section~\ref{sec:r2}. Detailed per-dataset analyses (why LoRA-Gen fails to improve GA, the HotpotQA evaluator collapse, and per-dataset LoRA-Both calibration) are given in Appendix~\ref{sec:lora_detail}.

\paragraph{Takeaway.}
Generation SFT suppresses evaluation discrimination (acquiescence, $\mathrm{ER} \to 100\%$); evaluation SFT suppresses generation and can destroy the evaluator on individual datasets. Per-dataset LoRA-Both is the only configuration achieving $|\Delta| \leq 1$\,pp without acquiescence on three of four datasets---but cannot recover base GA on harder datasets. The generation--evaluation asymmetry is an intrinsic property of in-context QA that SFT can redistribute but not eliminate.

\section{Conclusion}
\label{sec:conclusion}

Across four benchmarks and two models, generation accuracy exceeds self-evaluation on three of four---self-evaluation is not uniformly easier---with MuSiQue the exception due to a generation-difficulty ceiling at high hop counts. Attention analysis reveals a consistent structural cause: the evaluation task attends to context 3--5x less than generation does and barely reads the candidate answer. LoRA fine-tuning confirms this asymmetry is not a training artifact: generation fine-tuning induces over-acceptance and evaluation fine-tuning degrades generation.

Evaluation failures in self-assessment pipelines are structurally rooted in the model's attention-routing strategy at inference time, not in insufficient training signal. Pipelines relying on self-assessment---especially on multi-hop or numerical tasks---should account for the direction and magnitude of the $\Delta$ gap reported here, and treat acquiescence bias as a real risk whenever generation fine-tuning is part of the training recipe.

\section*{Limitations}


The attention analysis is restricted to the open-source Llama-3.1-8B-Instruct; GPT-4o-mini's internals are inaccessible. We also analyze only the first-token decision point---i.e.\ the attention distribution at the position predicting the first output token. For short-answer QA this is the most consequential position (gold answers are typically 1--3 tokens, and $\mathcal{T}_\text{eval}$'s output is exactly one token), but the analysis does not generalize directly to long-form generation, where later positions could exhibit qualitatively different attention patterns.

Our mechanistic account is most informative for the three benchmarks where $\Delta < 0$ (SQuAD~2.0, DROP, HotpotQA); we do not isolate a corresponding mechanism for the positive $\Delta$ on MuSiQue. The MuSiQue inversion is more parsimoniously explained by generation difficulty reaching a floor at high hop counts (Section~\ref{sec:r2}) than by any attention-level advantage in $\mathcal{T}_\text{eval}$ specific to that dataset. The fact that the model barely attends to the candidate answer is itself surprising and a target for follow-up: it suggests that decisions on $\mathcal{T}_\text{eval}$ may often be driven by structural priors rather than by a genuine semantic comparison. Locating the layer at which the candidate slot fuses with the context (e.g.\ via logit-lens or tuned-lens analysis) is a natural next step we leave to future work.

All prompts are fixed and zero-shot; results may differ with few-shot demonstrations or paraphrased prompts, and we leave prompt sensitivity analysis to future work. All benchmarks are English-only; cross-lingual generalization is not guaranteed. Our framework is scoped to QA settings where answer correctness admits an unambiguous binary label; abstractive or opinion-based QA presents a fundamentally different evaluation challenge outside our current scope.

\section*{Ethical Considerations}

This work studies the behavior of publicly available LLMs (Llama-3.1-8B-Instruct, GPT-4o-mini, GPT-4o) on existing English-language QA benchmarks (SQuAD~2.0, DROP, HotpotQA, MuSiQue), all of which are released for research use. No new data was collected, no human subjects were involved, and the findings are descriptive and analytical in nature. We do not foresee any direct harms arising from this research. There are no major ethical concerns with this submission.

\bibliography{custom}

\newpage

\appendix
\section*{Appendix}

\section{Reproducibility}
\label{sec:reproducibility}

\paragraph{General setup.}
All models are queried at temperature $\tau = 0$ with greedy decoding. API calls use fixed seeds where supported. Evaluation responses that do not begin with \textsc{Correct} or \textsc{Incorrect} after stripping leading whitespace are treated as abstentions and excluded from metric computation; abstention rates are reported alongside all results.

\paragraph{Attention analysis.}
We set \texttt{attn\_implementation="eager"} in the Llama-3.1-8B-Instruct forward pass to expose raw attention tensors. A single forward pass is performed per prompt; only the last-token attention row is retained to keep memory bounded. Context, question, and answer spans are located via the tokenizer's character-to-token offset mapping. Attention weights are averaged over all heads and over layers 24--31 (the final eight layers of the 32-layer model).

\paragraph{LoRA fine-tuning.}
We fine-tune Llama-3.1-8B-Instruct with LoRA adapters (rank 16, $\alpha = 32$) applied to \texttt{q\_proj}, \texttt{k\_proj}, \texttt{v\_proj}, and \texttt{o\_proj}. Training uses cross-entropy loss on response tokens only (\texttt{completion\_only\_loss=True}), 3 epochs, effective batch size 16, learning rate 2e-4 with a cosine schedule, and bfloat16 precision. Hard negatives are generated with up to 5 hallucination attempts per record (Section~\ref{sec:transferability}). After training, each adapter is evaluated on the same 500-instance validation split used for the task-asymmetry analysis.

\paragraph{Hardware.}
All API-based experiments (oracle scoring, GPT-4o-mini generation and evaluation, oracle reliability study) are run from a standard CPU host. Llama-3.1-8B-Instruct inference, the attention analysis, and all LoRA training and evaluation are run on a server with $4\times$ NVIDIA A100-SXM4-80GB GPUs (CUDA 13.0).

\section{Use of Datasets}
\label{sec:dataset_usage}

This appendix documents the per-experiment sampling choices and the rationale behind them. Table~\ref{tab:datasets_extended} lists the full official split sizes, the per-experiment sample counts, and the average context length per dataset.

\begin{table*}[h]
\centering
\resizebox{\textwidth}{!}{%
\begin{tabular}{llrrrrr}
\toprule
& & \multicolumn{2}{c}{\textbf{Full splits}} & \multicolumn{3}{c}{\textbf{Used per experiment}} \\
\cmidrule(lr){3-4}\cmidrule(lr){5-7}
\textbf{Dataset} & \textbf{Type} & \textbf{\#Train} & \textbf{\#Val (filtered)} & \textbf{LoRA train} & \textbf{Task-asym.\,/\,LoRA eval} & \textbf{Attention (verified)} \\
\midrule
SQuAD~2.0 & Extractive & 130,319 & 5,928 & 5,000 & 500 & 49 \\
DROP      & Numerical  &  77,409 & 5,889 & 5,000 & 500 & 46 \\
HotpotQA  & 2-hop      &  90,447 & 6,947 & 5,000 & 500 & 44 \\
MuSiQue   & 2--4-hop   &  19,938 & 2,417 & 5,000 & 500 & 45 \\
\bottomrule
\end{tabular}}
\caption{Extended split usage. \#Train is the official training split size; \#Val (filtered) is the validation size after the dataset-specific filters described in §\ref{sec:experiments} (e.g., answerable only for SQuAD~2.0, number-type only for DROP). \textbf{LoRA train}: per-dataset budget drawn from the train split for LoRA-Gen, LoRA-Eval, and LoRA-Both. \textbf{Task-asym.\,/\,LoRA eval}: shared evaluation sample, drawn once and reused across the base model and all LoRA checkpoints for direct comparability. \textbf{Attention (verified)}: subset of the 50-per-dataset oracle-reliability sample where GPT-4o and GPT-5.4 agree, used as the basis for the attention analysis (184 total). MuSiQue validation hop distribution: 1,252 / 760 / 405 for 2 / 3 / 4 hops.}
\label{tab:datasets_extended}
\end{table*}

\paragraph{Cost and compute.}
The most expensive operation per evaluation instance is the oracle call (GPT-4o), invoked once for every $(c, q, a, a^*)$ tuple. Across two models (Llama-3.1-8B-Instruct and GPT-4o-mini) and four datasets, the full filtered validation sets would total roughly 42,000 oracle-bearing instances, against $\sim$4,000 at 500 per (model, dataset). Each task-asymmetry instance further triggers a generation call and a self-evaluation call, tripling the API budget. For the attention analysis the binding constraint is GPU memory rather than API cost: each forward pass with \texttt{output\_attentions=True} requires $O(L \cdot H \cdot S^2)$ activations, which for MuSiQue's $\sim$2--4\,k-token contexts is already GPU-heavy on an A100. The LoRA fine-tuning runs are similarly compute-bound---three independent adapter trainings, 3 epochs each at effective batch 16, total $\sim$30 GPU-hours even at the 5,000-per-dataset training budget.

\paragraph{Statistical sufficiency for a binary outcome.}
GA, EA, EP, ER, and $\Delta$ are all binomial-proportion quantities. At $n = 500$ the 95\% confidence half-width is at most $\pm 4.4$\,pp (at $p = 0.5$) and tighter at the observed values (e.g.\ $\pm 2.4$\,pp at $\mathrm{GA} = 95\%$). The $\Delta$ values reported in Table~\ref{tab:main_results} range from $-14.2$ to $+4.0$\,pp; on MuSiQue stratified by hop count (Table~\ref{tab:hop_analysis}) they reach $+10.7$\,pp---all well outside the noise floor of a 500-instance sample. Even the smallest cell in the hop analysis (MuSiQue 4-hop, $n = 78$) has a half-width of $\pm 11$\,pp, still narrower than its observed $\Delta = +7.7$. Increasing $n$ beyond 500 would refine the precision of individual cells but would not change the sign or qualitative pattern of any reported $\Delta$. Binary-outcome inference is unusually efficient in this respect: in contrast to continuous metrics such as ROUGE or BLEU where 500 samples are often borderline, accuracy-style metrics over Bernoulli outcomes are well-resolved at this scale.

\paragraph{Comparability across models, datasets, and conditions.}
Using a fixed seed and the same 500-instance sample per dataset across all conditions gives every model and every checkpoint exactly the same evaluation set. Two consequences follow. First, comparisons across models (Llama vs.\ GPT-4o-mini) and across LoRA checkpoints (Base vs.\ LoRA-Gen / LoRA-Eval / LoRA-Both) isolate the variable of interest from sampling variance---if a metric moves, it moved because of the model or the training, not because the underlying instance pool changed. Second, datasets carry equal weight in cross-dataset averages: a free-floating evaluation on full validation sets would let HotpotQA (6,947 filtered) and SQuAD~2.0 (5,928) dominate MuSiQue (2,417) when reporting any aggregate.

\paragraph{LoRA training budget.}
The 5,000-per-dataset training budget (20,000 total for LoRA-Gen, 40,000 for LoRA-Eval after hard negatives, 40,000 for LoRA-Both) serves the same comparability goal: it balances the four datasets in joint training, preventing the much larger SQuAD~2.0 and HotpotQA training pools from dwarfing MuSiQue. It also sits comfortably within the data regime where rank-16 LoRA adapters reliably saturate, so the choice trades off little expected adapter quality for substantial training-cost savings.

\paragraph{Attention-analysis sub-sample.}
The further restriction to the 184 jointly verified samples (where GPT-4o and GPT-5.4 agree) is a \textit{methodological} choice rather than a budgetary one. The attention analysis interpretation depends on knowing whether $\mathcal{T}_\text{gen}$'s output is genuinely correct and $\mathcal{T}_\text{eval}$'s judgment genuinely right or wrong; restricting to instances where two strong oracles agree removes ambiguity introduced by single-oracle noise. The per-dataset verified counts (49, 46, 44, 45) yield per-cell standard errors of $\sim$2--3\,pp on the attention-fraction estimates---comfortably tighter than the 9--15\,pp $\Delta_\text{ctx}$ effects we report in Section~\ref{sec:r3}.

\section{Extended Attention and Causal Ablation Analyses}
\label{sec:extended_attention}

This appendix reports three additional analyses that decompose the last-token attention findings of Section~\ref{sec:r3} and provide a more granular mechanistic account of why $\Delta < 0$ on three of the four benchmarks. The analyses are: (i)~a paragraph-level decomposition of context attention into gold-supporting vs.\ distractor spans (Appendix~\ref{sec:ext_selective}); (ii)~attention to context tokens that mention the candidate answer (Appendix~\ref{sec:ext_mentions}); and (iii)~causal ablations of the candidate-answer slot in $\mathcal{T}_\text{eval}$ (Appendix~\ref{sec:ext_causal}). All experiments use the same Llama-3.1-8B-Instruct model and analysis configuration (layers~24--31, head-averaged, last-token attention row) as Section~\ref{sec:r3}. Analyses (i) and (ii) reuse the 184 jointly verified samples; (iii)~is run on the full 500-instance subset per dataset for statistical power on accuracy estimates. Across the three analyses, MuSiQue's positive $\Delta$ is consistently the case the experiments are \emph{not} sufficient to explain; we revisit this limitation at the end of Appendix~\ref{sec:ext_causal}.

\subsection{Paragraph-level selectivity: supporting vs.\ distractor}
\label{sec:ext_selective}

\paragraph{Motivation.}
Section~\ref{sec:r3} reports that $\mathcal{T}_\text{eval}$ directs only 1.4--5.4\% of last-token attention at the context---a 3--5$\times$ reduction relative to $\mathcal{T}_\text{gen}$. A natural hypothesis is that this smaller budget is more \emph{selectively} concentrated on the gold supporting paragraphs (with the candidate answer acting as a retrieval anchor), which would be especially advantageous on MuSiQue, where retrieval cost is highest (2 of 20 paragraphs are gold).

\paragraph{Method.}
For HotpotQA and MuSiQue we re-fetch the per-paragraph \texttt{is\_supporting} annotations from the source HuggingFace datasets, split the concatenated context into paragraphs by the join delimiter used in preprocessing, and sum the head-averaged last-token attention mass into each paragraph's token range. Per-token concentration normalizes for paragraph length: $\text{sup\_per\_tok}$ is the mass on supporting paragraphs divided by the number of supporting-paragraph tokens, analogously for distractors.

\begin{table}[h]
\centering
\small
\begin{tabular}{llrrr}
\toprule
\textbf{Dataset} & \textbf{Task} & \textbf{n} & \textbf{sup share} & \textbf{ratio} \\
\midrule
\multirow{2}{*}{HotpotQA}
  & $\mathcal{T}_\text{gen}$  & 44 & 0.410 & 4.59 \\
  & $\mathcal{T}_\text{eval}$ & 44 & 0.188 & 1.38 \\
\midrule
\multirow{2}{*}{MuSiQue}
  & $\mathcal{T}_\text{gen}$  & 45 & 0.306 & 3.58 \\
  & $\mathcal{T}_\text{eval}$ & 45 & 0.155 & 1.21 \\
\bottomrule
\end{tabular}
\caption{Paragraph-level selectivity of last-token attention. \textit{sup share} is the fraction of paragraph-level context attention falling on gold supporting paragraphs (uniform-attention baseline: 0.20 for HotpotQA's 2-of-10, 0.10 for MuSiQue's 2-of-20). \textit{ratio} is the per-token attention concentration on supporting paragraphs divided by the same quantity on distractors; a value of 1 indicates no per-token preference. $\mathcal{T}_\text{gen}$ is sharply selective (3.6--4.6$\times$); $\mathcal{T}_\text{eval}$ is essentially uniform (1.2--1.4$\times$).}
\label{tab:selective}
\end{table}

\paragraph{Result.}
Table~\ref{tab:selective} reports the per-token concentration ratio. $\mathcal{T}_\text{gen}$ allocates 3.6--4.6$\times$ more attention per token to supporting paragraphs than to distractors. $\mathcal{T}_\text{eval}$ collapses this ratio to 1.2--1.4$\times$---essentially uniform across the context. The supporting share itself drops similarly: from 0.41 to 0.19 on HotpotQA, from 0.31 to 0.16 on MuSiQue (both eval values barely above the respective uniform baselines of 0.20 and 0.10).

\paragraph{Interpretation.}
On HotpotQA the generator's 4.59$\times$ per-token preference for supporting paragraphs reflects the targeted retrieval needed to chain the two relevant passages. The evaluator's near-uniform ratio (1.38$\times$) shows that no equivalent re-retrieval happens at the verification decision point, providing a paragraph-level mechanism for HotpotQA's large negative $\Delta = -14.2$: the generator selectively reads what it needs to answer; the evaluator does not selectively read what it would need to verify. On MuSiQue, $\mathcal{T}_\text{eval}$ exhibits the same collapse in selectivity (1.21$\times$) but $\Delta > 0$, so the inversion is \emph{not} attributable to sharper context routing. As Section~\ref{sec:r2} argues, MuSiQue's inversion is more parsimoniously explained by generation accuracy bottoming out at higher hop counts.

\subsection{Answer-mention concentration in the context}
\label{sec:ext_mentions}

\paragraph{Motivation.}
A second hypothesis is that $\mathcal{T}_\text{eval}$ performs needle-in-haystack verification: locate tokens in the context that mention the candidate answer and concentrate attention there. The candidate's presence in the prompt should make this strategy especially natural for $\mathcal{T}_\text{eval}$ relative to $\mathcal{T}_\text{gen}$.

\paragraph{Method.}
For each prompt we identify token positions in the context that match content words ($\geq 3$ characters) of the candidate answer, case-insensitively and with non-alphanumeric word boundaries (with a fallback to shorter words for single-digit DROP answers). We then compare the per-token attention concentration on those positions to that of the remaining (non-mention) context tokens.

\begin{table}[h]
\centering
\small
\begin{tabular}{llrr}
\toprule
\textbf{Dataset} & \textbf{Task} & \textbf{n} & \textbf{mention / non-mention} \\
\midrule
\multirow{2}{*}{SQuAD~2.0}
  & $\mathcal{T}_\text{gen}$  & 49 & 13.03 \\
  & $\mathcal{T}_\text{eval}$ & 49 & 1.35 \\
\midrule
\multirow{2}{*}{DROP}
  & $\mathcal{T}_\text{gen}$  & 46 & 5.05 \\
  & $\mathcal{T}_\text{eval}$ & 46 & 1.03 \\
\midrule
\multirow{2}{*}{HotpotQA}
  & $\mathcal{T}_\text{gen}$  & 44 & 17.24 \\
  & $\mathcal{T}_\text{eval}$ & 44 & 1.86 \\
\midrule
\multirow{2}{*}{MuSiQue}
  & $\mathcal{T}_\text{gen}$  & 45 & 22.66 \\
  & $\mathcal{T}_\text{eval}$ & 45 & 2.12 \\
\bottomrule
\end{tabular}
\caption{Per-token attention concentration ratio between context tokens that match content words of the candidate answer (``mentions'') and the remainder of the context. A value of 1 indicates no per-token preference. $\mathcal{T}_\text{gen}$ concentrates sharply on answer-mention positions (5--23$\times$); $\mathcal{T}_\text{eval}$ is near-uniform (1.0--2.1$\times$).}
\label{tab:mentions}
\end{table}

\begin{figure}[h]
\centering
\includegraphics[width=\linewidth]{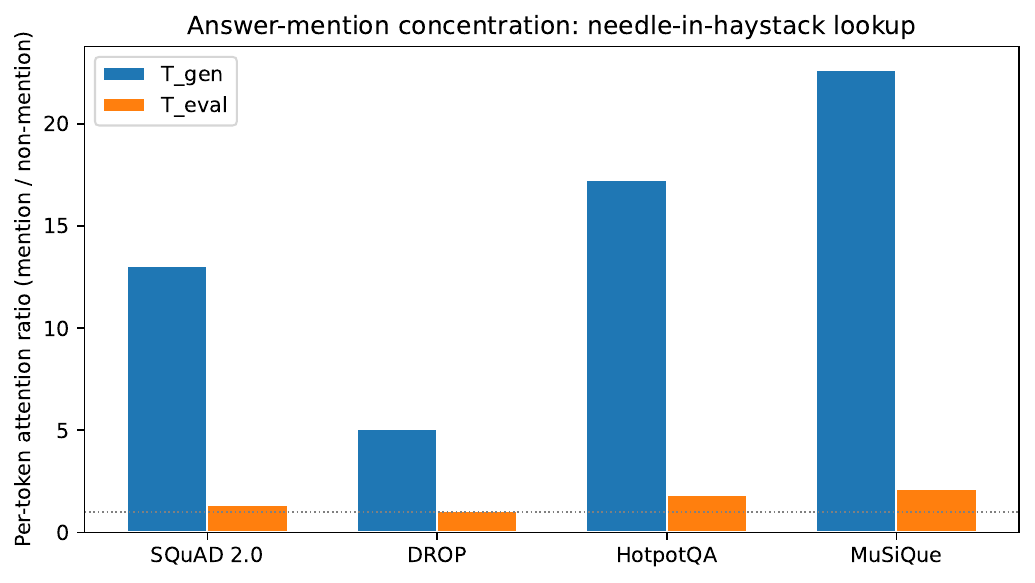}
\caption{Per-token attention ratio on answer-mention tokens vs.\ non-mention context tokens. $\mathcal{T}_\text{gen}$ exhibits sharp needle-in-haystack lookup behaviour; $\mathcal{T}_\text{eval}$ does not.}
\label{fig:mention_ratio}
\end{figure}

\paragraph{Result.}
Table~\ref{tab:mentions} and Figure~\ref{fig:mention_ratio} show that $\mathcal{T}_\text{gen}$ concentrates 5--23$\times$ more attention per token on answer-mention positions than on the rest of the context, with the ratio largest on the multi-hop benchmarks (HotpotQA 17.24, MuSiQue 22.66) where the answer span must be located within long, distractor-heavy contexts. $\mathcal{T}_\text{eval}$ collapses this ratio to 1.0--2.1$\times$.

\paragraph{Interpretation.}
$\mathcal{T}_\text{gen}$, not $\mathcal{T}_\text{eval}$, is the task that performs the answer-locating lookup. This asymmetry provides a complementary mechanistic explanation for $\Delta < 0$ on all three benchmarks where the gap is negative: the generator's sharp lookup converts context engagement into a correct span---most strongly on the multi-hop HotpotQA (17.24$\times$, $\Delta = -14.2$) and the extractive SQuAD~2.0 (13.03$\times$, $\Delta = -3.4$), with a milder but qualitatively identical pattern on DROP (5.05$\times$, $\Delta = -1.0$)---while the evaluator's near-uniform attention leaves it unable to perform the same lookup at verification time, so subtly-correct answers are misjudged as wrong (and vice versa) more often than the generator misses them in the first place. On MuSiQue the eval-side ratio is similarly diffuse (2.12$\times$) yet $\Delta > 0$, again indicating that the inversion is not produced by the attention pattern at the last-token decision and is instead consistent with the generation-floor account of Section~\ref{sec:r2}. Together with Appendix~\ref{sec:ext_selective}, the picture across negative-$\Delta$ datasets is consistent: $\mathcal{T}_\text{eval}$'s last-token attention is \emph{structurally diffuse}, not selectively targeted at task-relevant content, and this diffuseness is what costs the evaluator the accuracy that synthesis enjoys.

\subsection{Causal ablations of the candidate-answer slot}
\label{sec:ext_causal}

\paragraph{Motivation.}
The analyses above are correlational. To probe \emph{causally} how much of $\mathcal{T}_\text{eval}$'s behavior is driven by the candidate answer~$a$ in the prompt, we intervene on the candidate-answer slot in two complementary ways.

\begin{itemize}[noitemsep, topsep=2pt]
  \item \textbf{C-MASK} replaces $a$ with the placeholder \texttt{[REDACTED]}. The original oracle verdict remains the reference label, so EA under C-MASK measures how often the model recovers the right answer \emph{without} access to the candidate.
  \item \textbf{C-SWAP} replaces $a$ with the gold answer of another instance from the same dataset, selected by a fixed index offset for reproducibility (swap-collisions with the original gold are dropped). The new candidate is almost always wrong, so the reference label is \textsc{Incorrect}; the rejection rate measures whether $\mathcal{T}_\text{eval}$ is performing real verification or merely rubber-stamping the candidate.
\end{itemize}

Both ablations are run on all 500 instances per dataset; the binary judgment is recovered from a single forward pass by comparing the logits of the first BPE tokens of ``Correct'' and ``Incorrect''.

\begin{table}[h]
\centering
\resizebox{\columnwidth}{!}{%
\begin{tabular}{lrrrrr}
\toprule
\textbf{Dataset} & \textbf{n} & \textbf{EA base} & \textbf{EA C-MASK} & \textbf{$\Delta$} & \textbf{flip} \\
\midrule
SQuAD~2.0 & 500 & 92.2 & 68.8 & $-23.4$ & 31.0 \\
DROP      & 500 & 62.4 & 56.6 & $-5.8$  & 38.6 \\
HotpotQA  & 500 & 69.0 & 49.4 & $-19.6$ & 37.2 \\
MuSiQue   & 500 & 59.2 & 50.0 & $-9.2$  & 36.4 \\
\bottomrule
\end{tabular}}
\caption{C-MASK ablation: candidate $a$ replaced by \texttt{[REDACTED]}. $\Delta = \mathrm{EA}_\text{C-MASK} - \mathrm{EA}_\text{base}$ (all four are drops). \textbf{flip}: fraction of instances where the C-MASK judgment differs from baseline. MuSiQue's EA collapses to exactly chance (50.0\%).}
\label{tab:cmask}
\end{table}

\begin{table}[h]
\centering
\small
\begin{tabular}{lrrr}
\toprule
\textbf{Dataset} & \textbf{n} & \textbf{Rej.\%} & \textbf{Rej.\% (base=\textsc{Cor.})} \\
\midrule
SQuAD~2.0 & 500 & 99.0 & 98.9 \,(n\,=\,467) \\
DROP      & 492 & 95.7 & 95.4 \,(n\,=\,259) \\
HotpotQA  & 500 & 99.6 & 99.4 \,(n\,=\,349) \\
MuSiQue   & 499 & 99.8 & \textbf{100.0} \,(n\,=\,211) \\
\bottomrule
\end{tabular}
\caption{C-SWAP ablation: candidate $a$ replaced by another instance's gold answer (same dataset; 9 collisions with the original gold were dropped). Reference label is \textsc{Incorrect}; \textbf{Rej.\%} is the fraction of judgments that correctly say \textsc{Incorrect}. The final column conditions on the baseline judgment being \textsc{Correct}, isolating cases where the model flipped from endorsement to rejection. C-SWAP rejection on MuSiQue is universal under this conditioning.}
\label{tab:cswap}
\end{table}

\paragraph{Result.}
C-MASK (Table~\ref{tab:cmask}) drops EA by 5.8--23.4\,pp and flips 31--39\% of judgments; the candidate-answer slot is materially load-bearing on every dataset. MuSiQue's EA collapses from $59.2$\% to chance (50.0\%) under C-MASK---the strongest single signal of candidate dependence in the experiment. C-SWAP (Table~\ref{tab:cswap}) rejects the swapped (incorrect) candidate on 95.7--99.8\% of instances, and \emph{universally} (100\%) on the MuSiQue subset where the baseline had said \textsc{Correct}.

\paragraph{Interpretation.}
The two probes are mutually supportive. C-SWAP shows that $\mathcal{T}_\text{eval}$ \emph{is} performing genuine verification when the candidate is unambiguously wrong (it does not rubber-stamp), and C-MASK shows that the candidate slot is also \emph{required}---when it is removed, EA degrades sharply on every dataset. The evaluator's strategy is therefore best characterized as \textit{candidate-anchored shallow verification}: a fast check anchored at the candidate-answer slot in the prompt rather than at the last-token attention row over the context. This account directly explains the consistent pattern of $\Delta < 0$ on SQuAD~2.0, DROP, and HotpotQA: when the generator produces a near-miss, the evaluator's shallow check is insufficient to detect the subtle error---even though, as C-SWAP confirms, it would readily reject an unrelated answer. Combined with Appendices~\ref{sec:ext_selective}--\ref{sec:ext_mentions}, the mechanistic picture across these three datasets is consistent: verification is integrated into the residual stream by earlier layers around the candidate-answer slot, after which the last-token attention serves mainly to read out a decision that has already been formed, with the evaluator paying for that efficiency in lost accuracy relative to the more attention-engaged generator.

\paragraph{Note on MuSiQue.}
The eval-side attention pattern on MuSiQue is qualitatively the same as on the negative-$\Delta$ datasets (diffuse paragraph attention, low mention concentration), the C-MASK drop is in the same direction (post-ablation EA reaches exactly chance), and C-SWAP rejection is similarly near-universal. The three analyses thus do not isolate a MuSiQue-specific mechanism; we discuss this limitation and the most plausible alternative account in the Limitations section.

\subsection{Additional multi-hop dataset: 2WikiMultiHopQA}
\label{sec:ext_2wiki}

\paragraph{Motivation.}
The four benchmarks of the main analysis include two multi-hop datasets (HotpotQA, MuSiQue) that already display divergent behavior: $\Delta < 0$ on HotpotQA but $\Delta > 0$ on MuSiQue. To test whether the mechanistic findings of Section~\ref{sec:r3} generalize beyond these two datasets, we replicate the task-asymmetry and attention analyses on a third multi-hop benchmark, 2WikiMultiHopQA~\citep{ho-etal-2020-constructing}, which has the additional benefit of providing human-annotated question-type labels (\textit{bridge-comparison}, \textit{comparison}, \textit{compositional}, \textit{inference}). This enables a within-dataset breakdown of how reasoning structure modulates the gap and the underlying attention pattern.

\paragraph{Method.}
We sample 500 validation instances from 2WikiMultiHopQA following the same protocol used in Section~\ref{sec:experiments}, run Llama-3.1-8B-Instruct as $\mathcal{L}$ for both $\mathcal{T}_\text{gen}$ and $\mathcal{T}_\text{eval}$ at $\tau = 0$, and score the generations with GPT-4o as the oracle $\mathcal{L}^*$. The attention analysis follows Appendix~\ref{sec:extended_attention}: \texttt{attn\_implementation="eager"}, last-token attention row, head-averaged, restricted to layers 24--31. We use all $n = 495$ valid instances (those with a non-error oracle verdict) rather than the GPT-4o\,$\cap$\,GPT-5.4 verified subset; the GPT-5.4 endpoint used in Section~\ref{sec:oracle_reliability} was unavailable for this supplementary run.

\paragraph{Behavioral result.}
Table~\ref{tab:2wiki_behavioral} reports the task-asymmetry metrics. The overall $\Delta = -11.1$ on 2WikiMultiHopQA places this benchmark in the negative-$\Delta$ regime, qualitatively closer to HotpotQA ($-14.2$) than to MuSiQue ($+3.6$). However, the within-dataset breakdown reveals substantial heterogeneity: three of the four question types yield negative $\Delta$ (with bridge-comparison reaching $-25.2$), while the \textit{inference} subset---requiring composing a one-hop deduction over an explicitly stated relation---exhibits a strongly positive $\Delta = +21.6$. This mirrors the MuSiQue inversion within a single dataset and isolates it to a specific reasoning structure rather than a dataset-level artifact.

\begin{table}[h]
\centering
\small
\begin{tabular}{lrrrr}
\toprule
\textbf{Subset} & \textbf{n} & \textbf{GA} & \textbf{EA} & \textbf{$\Delta$} \\
\midrule
Overall                 & 495 & 67.9 & 56.8 & $-11.1$ \\
\midrule
bridge-comparison       & 111 & 70.3 & 45.0 & $-25.2$ \\
comparison              & 118 & 77.1 & 60.2 & $-16.9$ \\
compositional           & 215 & 70.7 & 62.3 & $-8.4$  \\
inference               &  51 & 29.4 & 51.0 & $+21.6$ \\
\bottomrule
\end{tabular}
\caption{2WikiMultiHopQA task asymmetry for Llama-3.1-8B-Instruct, overall and by question type. The \textit{inference} subset is the only one with $\Delta > 0$, replicating the MuSiQue-style inversion within a single multi-hop dataset.}
\label{tab:2wiki_behavioral}
\end{table}

\paragraph{Attention result.}
Table~\ref{tab:2wiki_attn} reports the last-token attention fractions. The aggregate pattern of Section~\ref{sec:r3} reproduces cleanly: $\mathcal{T}_\text{eval}$ allocates only 3.8\% of last-token attention to context, compared with 15.6\% for $\mathcal{T}_\text{gen}$---a $4.08\times$ drop, squarely within the 3--5$\times$ range reported on the four primary benchmarks. The candidate answer $a$ receives only $0.33$\% of attention overall (range $0.21$--$0.43$\% across question types), matching the $0.3$--$0.5$\% range from the main analysis. The context-attention de-allocation is consistent across all four question types ($-8.5$ to $-14.2$\,pp).

\begin{table}[h]
\centering
\resizebox{\columnwidth}{!}{%
\begin{tabular}{lrrrr}
\toprule
\textbf{Subset} & $\mathcal{T}_\text{gen}{\to}c$ & $\mathcal{T}_\text{eval}{\to}c$ & $\mathcal{T}_\text{eval}{\to}a$ & $\Delta_\text{ctx}$ \\
\midrule
Overall            & 15.6 & 3.8 & 0.33 & $-11.8$ \\
\midrule
bridge-comparison  & 13.9 & 3.9 & 0.31 & $-10.0$ \\
comparison         & 11.8 & 3.3 & 0.43 & $-8.5$  \\
compositional      & 17.9 & 4.0 & 0.32 & $-13.9$ \\
inference          & 18.3 & 4.1 & 0.21 & $-14.2$ \\
\bottomrule
\end{tabular}}
\caption{2WikiMultiHopQA last-token attention fractions (\%) directed to context $c$ and candidate answer $a$, averaged over layers 24--31 and all heads. $\Delta_\text{ctx}$ is the eval-minus-gen difference on context (in percentage points). The 3--5$\times$ context de-attention pattern of Section~\ref{sec:r3} reproduces uniformly across all four question types.}
\label{tab:2wiki_attn}
\end{table}

\paragraph{Interpretation.}
The structural attention signature of $\mathcal{T}_\text{eval}$---a 3--5$\times$ reduction in context attention coupled with negligible attention on the candidate answer---generalizes to a third multi-hop dataset, supporting its status as a general property of the evaluation task rather than a feature of any single benchmark. The within-dataset breakdown additionally suggests that the sign of $\Delta$ is governed by an interaction between this structural pattern (which is broadly invariant across reasoning types) and the difficulty of the generation task itself: on \textit{inference} questions, where Llama's generation accuracy collapses to $29.4$\%, evaluation becomes easier than generation despite the same attention de-allocation. This is consistent with the generation-floor account proposed for MuSiQue in Section~\ref{sec:r2}, and indicates that the same mechanism produces both signs of the gap depending on where generation accuracy lands.

\paragraph{Caveat and outlook.}
Two qualifications constrain the conclusions of this section. First, the $n = 495$ samples here are not filtered through a GPT-5.4 super-oracle (Section~\ref{sec:oracle_reliability}), so the verdicts that enter the analysis are slightly noisier than those of the main paper; we do not expect this to invert any sign, but exact magnitudes (especially in the small \textit{inference} subset, $n = 51$) should be treated as indicative. Second, this remains a single additional dataset and a single model, with one specific multi-hop ontology of question types; the bridge-comparison vs.\ inference dissociation observed here is suggestive but does not yet establish a typology. More broadly, the pattern emerging across our three multi-hop benchmarks---HotpotQA, MuSiQue, and 2WikiMultiHopQA---is that the mechanistic attention signature is robust, but the behavioral $\Delta$ depends sensitively on the underlying reasoning structure and on the model's generation ceiling for that structure. Disentangling these two factors---ideally with controlled variations of hop count, distractor density, and supporting-fact configuration on a common context substrate, and with stronger models---is a natural direction for future work.

\section{Detailed LoRA Transfer Analysis}
\label{sec:lora_detail}

This appendix provides the per-dataset analyses that underlie the summary results of Section~\ref{sec:r4}.

\paragraph{Why does LoRA-Gen not improve GA?}
Contrary to the pre-experiment expectation, LoRA-Gen \emph{lowers} GA on three of four datasets (SQuAD~2.0: $-1.6$, DROP: $-2.2$, MuSiQue: $-6.6$); only HotpotQA improves ($+0.8$). Three mechanisms plausibly contribute. (i) \textbf{Saturation}: Llama-3.1-8B's pretraining and instruction-tuning data very likely cover these benchmarks closely, leaving little headroom (SQuAD~2.0 starts at GA $= 95.6\%$). (ii) \textbf{Format shift}: cross-entropy on response tokens pushes the model toward the exact gold-string form, which can be more brittle to the oracle's paraphrase-tolerant scoring than the base model's pre-finetuning output. (iii) \textbf{Joint-training interference}: the four datasets span extractive, numerical, $2$-hop, and $2{-}4$-hop reasoning with very different response distributions; the rank-$16$ adapter must encode all four styles simultaneously, and MuSiQue---the most heterogeneous response distribution---shows the largest GA regression ($-6.6$\,pp), consistent with averaging across styles. Training loss converges smoothly on all three runs (final $1.29$, $1.16$, $1.16$ for LoRA-Gen, LoRA-Eval, LoRA-Both); validation loss, tracked with a 500-instance held-out split in the per-dataset experiments below, is monotonically decreasing across all twelve per-dataset adapters, ruling out overfitting.

\paragraph{Per-dataset adapters confirm over-acceptance is intrinsic, not interference-driven.}
To isolate mechanism~(iii), we train one adapter per dataset per task variant (twelve adapters total; Table~\ref{tab:transfer_main}, per-ds rows). If joint-training interference were the primary cause of GA regression, per-dataset LoRA-Gen should substantially recover GA. Instead, GA regressions persist on SQuAD~2.0 ($-3.2$\,pp vs.\ base), HotpotQA ($-\,0.2$\,pp), and MuSiQue ($-7.2$\,pp); only DROP improves ($+0.6$\,pp, $64.0$ vs.\ $63.4$ base), suggesting that dataset had genuine headroom that joint training was suppressing. Crucially, evaluator recall remains near $100\%$ under per-dataset LoRA-Gen on SQuAD~2.0 and HotpotQA, and the MuSiQue adapter---despite having ER\,$= 40.6\%$ and a conservative $13\%$ abstention rate---still shows a GA regression of $7.2$\,pp. Over-acceptance is therefore an intrinsic consequence of generation-only SFT rather than a multi-dataset averaging artifact.

\paragraph{Per-dataset LoRA-Eval exposes a HotpotQA evaluator collapse.}
Per-dataset LoRA-Eval degrades GA even more severely than the joint variant (MuSiQue: $36.8\%$ vs.\ $38.0\%$; DROP: $41.4\%$ vs.\ $46.6\%$), confirming that evaluation-only SFT universally suppresses generation. The most striking finding is HotpotQA: the per-dataset evaluator collapses to EA\,$= 47.5\%$ ($\Delta = -29.7$\,pp), well below the base model's $69.0\%$, despite GA remaining reasonable at $77.2\%$. The joint variant, trained on all four datasets, had EA\,$= 63.6\%$ on HotpotQA---worse than base but far less catastrophic. Isolating to HotpotQA therefore \emph{worsens} the evaluator on that very dataset, which is paradoxical if evaluation is task-specific. The dissociation---competent generation, broken evaluation---on the same model and dataset is consistent with $2$-hop reasoning drawing on representational resources that are shared between the two tasks; evaluation-only SFT disrupts that shared substrate without the counterbalance of a generation signal.

\paragraph{Per-dataset LoRA-Both is the only calibrated regime.}
Joint LoRA-Both already reduced GA regressions relative to LoRA-Gen, but its ER remained near $100\%$ (Table~\ref{tab:transfer_eper}), indicating persistent over-acceptance. Per-dataset LoRA-Both breaks this pattern: $|\Delta| \leq 1$\,pp on SQuAD~2.0 ($-0.2$), DROP ($-0.8$), and HotpotQA ($0.0$), with no ER collapse. MuSiQue retains $\Delta = +9.2$, consistent across all variants and attributable to the structural difficulty of multi-hop generation rather than training bias. However, even the most balanced training regime cannot recover the base GA: MuSiQue falls from $55.6$ to $44.3\%$, a $11.3$\,pp regression that survives both joint and per-dataset training. The generation--evaluation gap is therefore not an artifact that SFT can straightforwardly correct; it reflects an allocation of parametric capacity that fine-tuning reallocates but does not expand.

\section{Case Studies: Self-Evaluation Error Analysis}
\label{sec:case_studies}

We present four representative instances from the base Llama-3.1-8B-Instruct evaluation (SQuAD~2.0 and DROP) illustrating the four possible outcomes of the generation--self-evaluation pipeline. Each example shows the context $c$, question $q$, gold answer $a^*$, generated answer $a$, oracle verdict $y^*$ (Eq.~3), and self-evaluation judgment $y$ (Eq.~2). Key spans are \textbf{bolded}.

\paragraph{True Positive (TP) --- DROP.}
The model generates the correct answer and correctly judges it as correct.

\begin{quote}
\small
\textbf{Context:} \ldots{} Baltimore continued to suppress Tampa Bay in the second half, as Stover kicked a \textbf{42-yard} field goal in the third quarter to put Baltimore up 20--0. \ldots

\smallskip
\textbf{Question:} How many yards did Stover kick in the third?

\smallskip
\textbf{Gold ($a^*$):} 42 \qquad \textbf{Generated ($a$):} 42 yards.

\smallskip
\textbf{Oracle ($y^*$):} \textit{Correct} \qquad \textbf{Self-eval ($y$):} \textit{Correct} $\surd$
\end{quote}

The answer is explicitly stated in the context. The self-evaluator has no difficulty confirming a direct numerical match.

\paragraph{True Negative (TN) --- SQuAD~2.0.}
The model confuses two entities and the self-evaluator correctly catches the error.

\begin{quote}
\small
\textbf{Context:} \textbf{Gasquet} (1908) claimed that the Latin name \textit{atra mors} (Black Death) first appeared in modern times in 1631 in a book on Danish history by \textbf{J.I.\ Pontanus}: ``Vulgo \& ab effectu atram mortem vocatibant.'' The name spread through Scandinavia and then Germany, gradually becoming attached to the mid 14th-century epidemic as a proper name.

\smallskip
\textbf{Question:} Who allegedly coined the name Black Death?

\smallskip
\textbf{Gold ($a^*$):} J.I.\ Pontanus \qquad \textbf{Generated ($a$):} Gasquet

\smallskip
\textbf{Oracle ($y^*$):} \textit{Incorrect} \qquad \textbf{Self-eval ($y$):} \textit{Incorrect} $\surd$
\end{quote}

The model is misled by sentence position: \textit{Gasquet} is the grammatical subject of the opening clause while \textit{J.I.\ Pontanus}---the alleged coiner---appears only in a subordinate clause. The model returns the historian who \emph{cited} Pontanus rather than Pontanus himself. Crucially, the self-evaluator correctly identifies ``Gasquet'' $\neq$ ``J.I.\ Pontanus'' and flags the answer as incorrect.

\paragraph{False Positive (FP) --- SQuAD~2.0.}
The model returns a context-grounded but wrong answer; the self-evaluator fails to detect the error.

\begin{quote}
\small
\textbf{Context:} British imperialism often used the concept of \textbf{Terra nullius} (Latin expression which stems from \textbf{Roman law} meaning `empty land'). The country of Australia serves as a case study: British settlement was premised on \textit{terra nullius}, and its settlers considered it unused by its sparse Aboriginal inhabitants.

\smallskip
\textbf{Question:} What law justified British imperialism?

\smallskip
\textbf{Gold ($a^*$):} Terra nullius \qquad \textbf{Generated ($a$):} Roman law.

\smallskip
\textbf{Oracle ($y^*$):} \textit{Incorrect} \qquad \textbf{Self-eval ($y$):} \textit{Correct} $\times$
\end{quote}

The model confuses the \emph{origin} of the legal concept (``stems from Roman law'') with the concept itself (Terra nullius). Because ``Roman law'' appears verbatim in the context and is semantically adjacent to the correct answer, the self-evaluator accepts it as plausible. Detecting the error requires composing the relation ``$X$ stems from $Y$'' $\neq$ ``$X = Y$''---an inference the evaluator fails to make, illustrating how surface-level context grounding misleads self-evaluation.

\paragraph{False Negative (FN) --- SQuAD~2.0.}
The model generates a factually correct but concise answer; the self-evaluator over-penalises the omission of descriptive detail.

\begin{quote}
\small
\textbf{Context:} \ldots{} Washington left with a small party, picking up along the way \textbf{Jacob Van Braam} as an interpreter; \textbf{Christopher Gist}, a company surveyor working in the area; and a few \textbf{Mingo led by Tanaghrisson}.

\smallskip
\textbf{Question:} Who was added to the party as Washington went on his way?

\smallskip
\textbf{Gold ($a^*$):} Jacob Van Braam as an interpreter; Christopher Gist, a company surveyor working in the area; and a few Mingo led by Tanaghrisson

\textbf{Generated ($a$):} Jacob Van Braam, Christopher Gist, and a few Mingo led by Tanaghrisson.

\smallskip
\textbf{Oracle ($y^*$):} \textit{Correct} \qquad \textbf{Self-eval ($y$):} \textit{Incorrect} $\times$
\end{quote}

The generated answer names all three correct parties but omits the role descriptors (``as an interpreter'', ``a company surveyor working in the area''). The oracle correctly judges the core factual content as sufficient; the self-evaluator applies a stricter criterion and marks the answer wrong. This illustrates a systematic source of false negatives: the evaluator conflates \emph{completeness} with \emph{correctness}, penalising concise but accurate answers.

\end{document}